\documentclass[10pt,twocolumn,letterpaper]{article}

\usepackage[pagenumbers]{cvpr} 

\usepackage{graphicx}
\usepackage{amsmath}
\usepackage{amssymb}
\usepackage{booktabs}
\usepackage{mathtools}
\usepackage{tabu}

\usepackage[pagebackref,breaklinks,colorlinks]{hyperref}

\usepackage[capitalize]{cleveref}
\crefname{section}{Sec.}{Secs.}
\Crefname{section}{Section}{Sections}
\Crefname{table}{Table}{Tables}
\crefname{table}{Tab.}{Tabs.}

\newcommand{\sref}[1]{Sec.~\ref{#1}}
\newcommand{\figref}[1]{Fig.~\ref{#1}}

\newcommand{\figureref}[1]{Figure~\ref{#1}}
\newcommand{\tableref}[1]{Table~\ref{#1}}

\begin{document}

\title{Unsupervised Change Detection Based on Image Reconstruction Loss}

\author{Hyeoncheol Noh \thanks{equal contribution}\\
Hanbat National University\\
{\tt\small hyeoncheol.noh@edu.hanbat.ac.kr}
\and
Jingi Ju \footnotemark[1]\\
Hanbat National University\\
{\tt\small jingi.ju@edu.hanbat.ac.kr}
\and
Minseok Seo \footnotemark[1]\\
Hanbat National University\\
{\tt\small minseok.seo@hanbat.ac.kr}
\and
Jongchan Park\\
Lunit Inc\\
{\tt\small jcpark@lunit.io}
\and
Dong-Geol Choi\\
Hanbat National University\\
{\tt\small dgchoi@hanbat.ac.kr}
}

\maketitle

\begin{abstract}
To train the change detector, bi-temporal images taken at different times in the same area are used. However, collecting labeled bi-temporal images is expensive and time consuming. To solve this problem, various unsupervised change detection methods have been proposed, but they still require unlabeled bi-temporal images. 
In this paper, we propose unsupervised change detection based on image reconstruction loss using only unlabeled single temporal single image. 
The image reconstruction model is trained to reconstruct the original source image by receiving the source image and the photometrically transformed source image as a pair. 
During inference, the model receives bi-temporal images as the input, and tries to reconstruct one of the inputs. The changed region between bi-temporal images shows high reconstruction loss.
Our change detector showed significant performance in various change detection benchmark datasets even though only a single temporal single source image was used. 
The code and trained models will be publicly available for reproducibility.

\end{abstract}

\section{Introduction}
\label{sec:intro}
In earth vision, change detection is a task to detect semantic changes in two high spatial resolution (HSR) images from different times and the same area (i.e. bi-temporal images).
Change detection is a very important task in the field of earth vision used for urban expansion, urban planning, environmental monitoring, and disaster assessment~\cite{hussain2013change,zhang2017separate}.

However, the manual comparison and change detection between two HSR images is a very labor-intensive and costly job.
To solve this problem, recent deep learning-based change detection methods~\cite{chen2021remote,fang2021snunet} have been proposed and the results are promising.
Due to the data-driven nature of deep learning methods, a large-scale training dataset of bi-temporal images and corresponding change labels is essential for supervised approaches~\cite{chen2021remote,fang2021snunet}.
%
%
The challenge lies in the expensive dataset: Collecting the correctly registered bi-temporal HSR images is expensive, and annotating the changes between them is more costly than general semantic segmentation~\cite{waqas2019isaid} or object detection datasets~\cite{xia2018dota}.
Another challenge is the imbalanced dataset: The change detection dataset requires two images taken at different times in the same area, and in real-world scenarios, changes are rare, so it is more difficult to collect a change detection dataset in which changes exist (e.g. class balanced).
%
%

%
\begin{figure}[t]
  \centering
  \includegraphics[width=\linewidth]{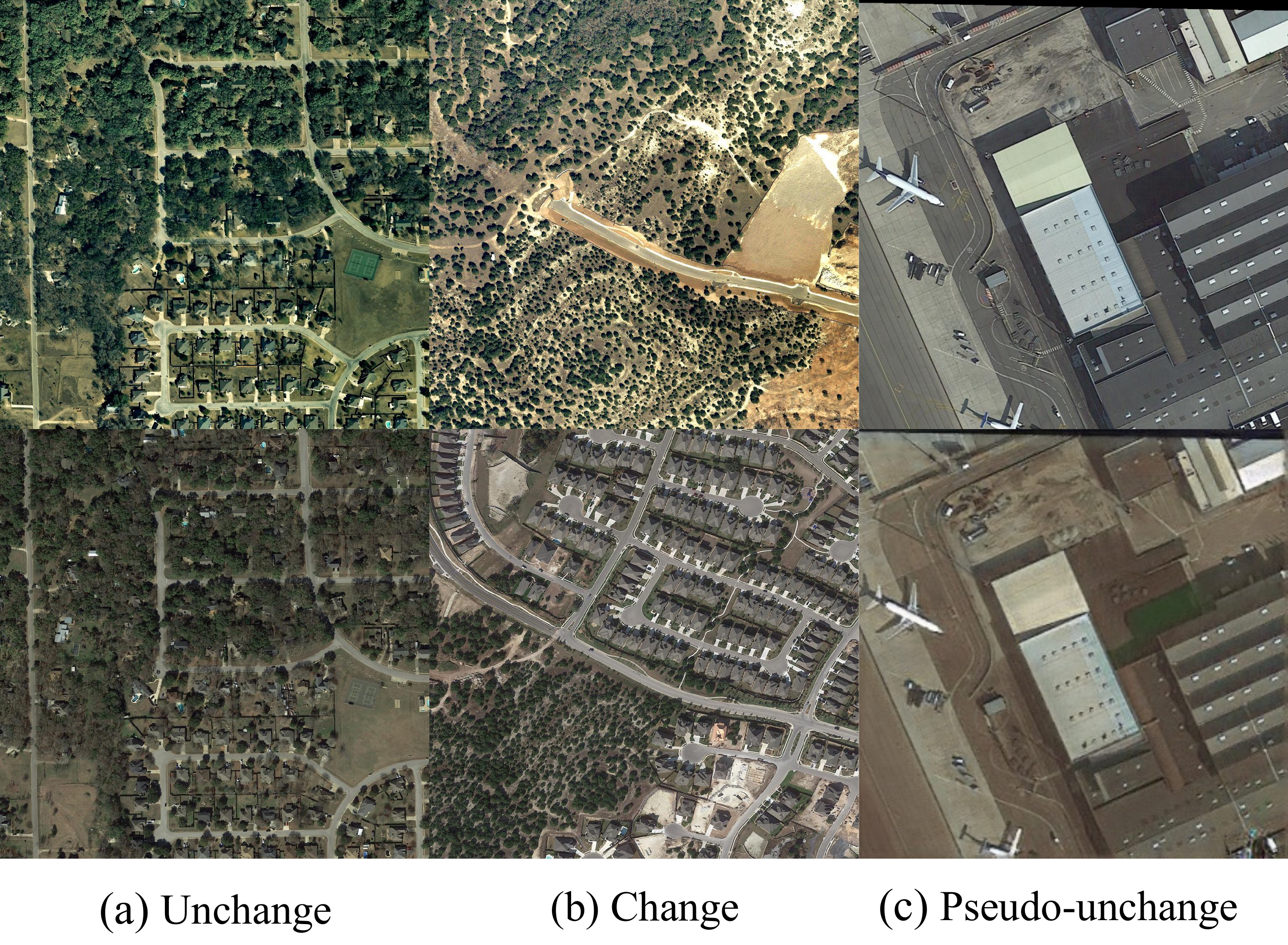}
  \caption{Qualitative comparison of unchanged bi-temporal pair images and changed bi-temporal pair images and pseudo-unchanged pair images. (a) and (b) are actual bi-temporal pair images, (c) is a pseudo-unchanged pair created by photometric transform.}
  \label{fig:fig1}
\end{figure}
To solve this data collection problem, various unsupervised change detection (UCD) methods~\cite{he2021unsupervised, de2019unsupervised, ren2020unsupervised, li2019unsupervised} have been proposed.
UCD approaches effectively solve the problem of expensive annotations in change detection, but they still require correctly registered bi-temporal HSR images, or the performance was low compared to supervised learning methods. 

In the existing UCD setting, because it does not explicitly train change and unchange, the prediction result is noisy both when the change occurs and when the change does not occur.
To solve this problem, most UCD methods use post-processing. However, they do not disclose how to post-processing, or they are overfitting for specific situations. (e.g. pixels with small area are removed)

Inspired by the study of unsupervised anomaly detection~\cite{li2021cutpaste,gong2019memorizing}, we rethink the UCD setup.
In both change detection and anomaly detection, change/anomaly situations are rare in the real world. 
The unsupervised anomaly detection methods~\cite{li2021cutpaste,gong2019memorizing} train the image reconstruction model with only normal data,  and the model is fitted to the normal distribution. During inference, normal inputs will be well reconstructed, as they fall in the normal distribution; on the other hand, anomaly inputs will have high reconstruction error, as they fall outside the normal distribution. 
%
%
%
\textit{Can UCD use reconstruction errors like unsupervised anomaly detection?}
Since unchanged pairs can be generated synthetically, we can train an image reconstruction model that trains normal distributions.
For example, if the change detector is trained on the unchanged area by pairing $X^{t1}$ with itself, it can be trained on the unchanged area without $X^{t2}$ images or changed labels.

In this paper, we propose unsupervised change detection based on image reconstruction loss (CDRL) using only unlabeled single-temporal single source images. 
The proposed method explicitly solves the challenges of data collection in change detection, as it does not require expensive bi-temporal HSR images, expensive annotations, nor balanced datasets with sufficient changes.
%
CDRL is trained to reconstruct the original source image by receiving the source image and the photometrically transformed source image as a pair.
The purpose of photometric transformations is to create pseudo-unchanged pairs that mimic unchanged pairs, as shown in \figref{fig:fig1}-(a) and \figref{fig:fig1}-(c). In the unchanged pairs, there are no structural changes by definition, but only style changes or photometric changes. The pseudo-changed pairs can be used to train CDRL instead of changed pairs.
%
%
Similar to unsupervised anomaly detection, CDRL receives only (pseudo-)unchanged pair images during training and is trained to reconstruct the original source image, so if untrained cases (changed pair~\figref{fig:fig1}-(b)) are input during inference, the reconstruction loss is high for that area.

However, unlike existing unsupervised anomaly detection studies, change detection usually receives two images, so there are two major problems.
First, the image reconstruction models should be able to reconstruct the original source image by receiving two pair images.
Second, the reconstruction models should focus more on the structure information of the photometrically transformed source image.
To solve this problem, we propose an image reconstruction model using generative adversarial networks based on encoder-decoder.
%
CDRL consists of a shared encoder to extract features from each image, and a decoder to fuse the features from the two images for image reconstruction.
%
To pay attention to the structure information of the photometrically transformed source image, spatial attention was performed only on the photometrically transformed source image.

%
To validate the efficacy of our proposed CDRL, we evaluated it on LEVIR-CD~\cite{Chen2020} and WHU-CD~\cite{liu2020novel}. 
Even though CDRL does not use bi-temporal pairs or pre-trained weights, CDRL outperforms the existing UCD method using bi-temporal pairs and the UCD method using pre-trained weights by a large margin.

\begin{figure*}[!t]
  \centering
  \includegraphics[width=\linewidth]{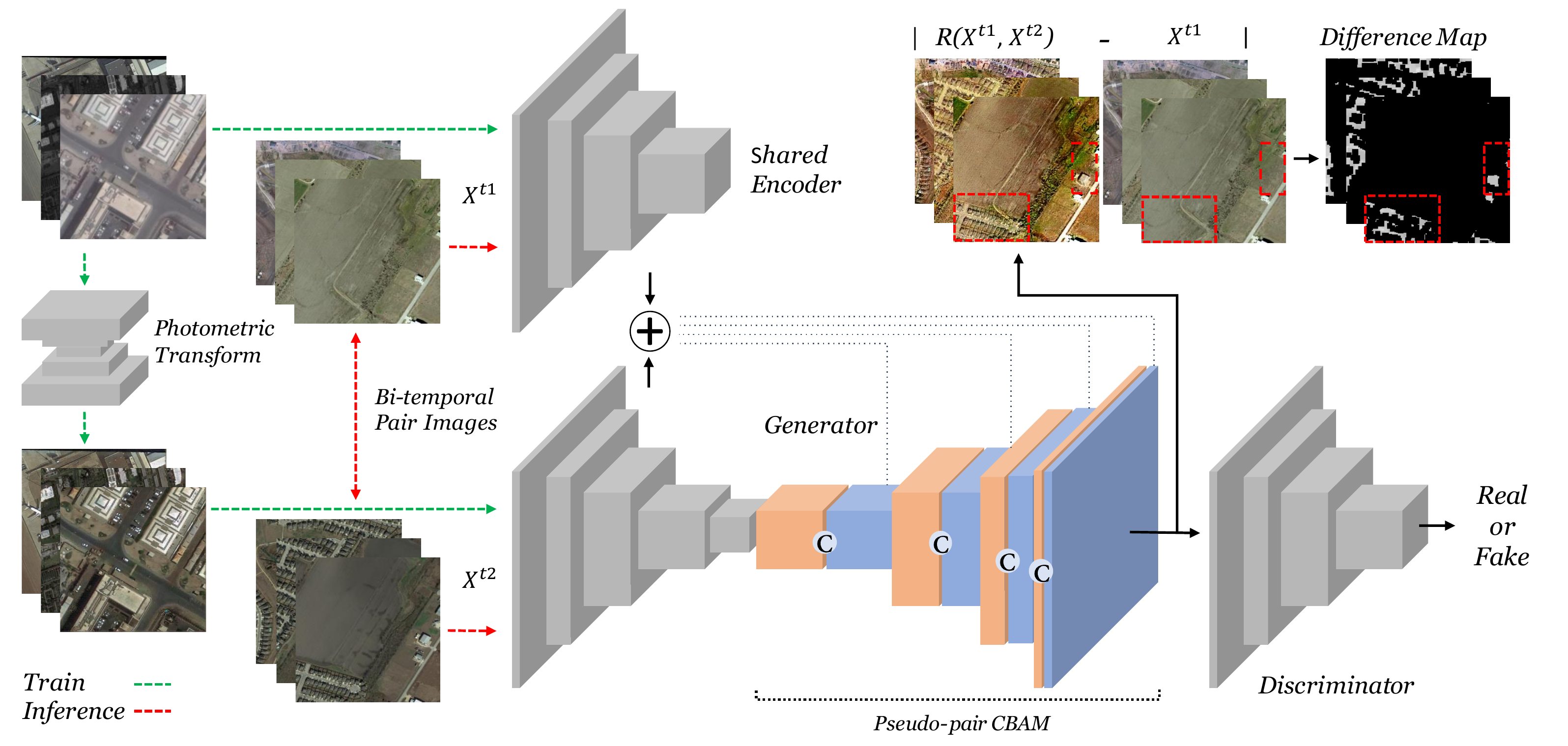}
  \caption{Overview of the overall framework of \textit{CDRL}. CDRL is trained to reconstruct $X^{t1}$ by receiving a pseudo-unchanged pair during training, and when a changed bi-temporal pair that is not learned during training is input during inference, the reconstruction loss is large in the region with large structure change.}
  \label{fig:fig2}
\end{figure*}

To sum up, our major contributions are as follows: 
\begin{itemize}
    \item We propose CDRL, a method to train a change detector on a single-temporal single source image in UCD. To the best of our knowledge, this is the first time to use a single-temporal single source image in UCD.
    \item We propose an encoder-decoder-based generative adversarial network that receives paired images as input.
    \item We evaluate CDRL on various change detection datasets, and CDRL outperforms previous UCD methods by a large margin.
\end{itemize}

\section{Related Work}
\label{sec:related}
The problem we are trying to solve is \textcircled{1} obtaining a matched bi-temporal image including a changed area is more difficult than obtaining a general single-temporal image \textcircled{2} pairwise annotation is very expensive and time consuming.
Therefore, this section focuses on the problems encountered by existing change detections, and finally briefly introduces the field of anomaly detection that we have been inspired by.

\subsection{Supervised Change Detection (SCD)}
Supervised change detection is largely divided into a method that uses only single-temporal information and an approach that performs temporal information modeling or different modeling~\cite{zheng2021change}.
A change detector that uses only single-temporal information, called post-classification comparison (PCC), trains a semantic segmentation model during training~\cite{zheng2021change, zheng2020foreground}.
After that, the semantic segmentation model predicts the change area through the xor operation of the results obtained by predicting images from two different times during inference.
PCC has the great advantage of not requiring coregistrated pair images, but this method only simply treats the change detection task as the semantic segmentation task and ignores the temporal information modeling, thus significantly decreasing the performance.

To solve this problem, change detection methods~\cite{chen2021remote,fang2021snunet} were proposed for temporal information modeling between pair images taken at different times in the same area. All of these methods achieved high performance, but the generalization performance of these models was not guaranteed because of the small size of the change detection benchmark datasets~\cite{lebedev2018change, tian2020hi, ji2018fully, chen2020spatial, daudt2019multitask, daudt2018urban, fujita2017damage, bourdis2011constrained, benedek2009change}.
The reason why the change detection benchmark datasets are small is that collecting bi-temporal pair images is much more difficult than collecting single-temporal images, and pairwise annotation is very expensive and time-consuming.

Since our proposed CDRL performs UCD using only unlabeled single-temporal single source images, it can alleviate the problem of collecting bi-temporal pair images and the cost and time-consuming problems of labeling.

\subsection{Unsupervised Change Detection (UCD)}
UCD is usually divided into a method~\cite{wu2013slow,thonfeld2016robust,blaschke2001object,im2008object} based on the concept of Change Vector Analysis~\cite{malila1980change} (CVA) or a method~\cite{ren2020unsupervised} based on a Generative Adversarial Network (GAN) using an unlabeled bi-temporal pair image.
However, because they use pre-trained weights without direct training on the dataset, the performance is low, or large-scale unlabeled bi-temporal pair images are required to train the GAN model.

Our proposed CDRL can be explicitly trained on an unchanged area and can be trained without bi-temporal pair images.
\subsection{Unsupervised Anomaly Detection}
The anomaly detection study we are inspired by is a reconstruction-based method~\cite{bergmann2019mvtec, perera2019ocgan}.
Reconstruction-based methods typically utilize generative models like auto-encoders or generative adversarial networks to encode and reconstruct the normal data. These methods hold the insights that the anomalies can not be reconstructed since they do not exist at the training samples.
These unsupervised anomaly detection methods achieved AUROC performance of over 95 in various benchmark datasets~\cite{bergmann2019mvtec}, even without explicitly training the anomaly data.

We also applied the fact that only unchanged pair (normal) images are trained during training like this reconstruction-based anomaly detection, and that when a changed pair (anomaly) is input during inference, the reconstruction loss is high.

\section{Method}
\label{sec:method}
This section describes the components of the CDRL in detail.
First, the training pipeline will be briefly described in ~\sref{sec:pipe}, and then, a method of performing photometric transform based on a single-temporal single source image will be described in ~\sref{sec:photo}.
~\sref{sec:model} describes the reconstructor that receives pair images and is trained as an objective to reconstruct the original source image.
Finally, ~\sref{sec:gan} describes the entire objective function of the CDRL including the GAN model.
\subsection{Overall Pipeline}
\label{sec:pipe}
CDRL performs photometric transform to create a pair image as a single-temporal single source image.
Photometric augmentation of simple rules such as brightness control and channel shuffling does not sufficiently express the style change of the corresponding bi-temporal pair image in the real-world unchanged area.
Therefore, in order to express the style change of the corresponding bi-temporal pair image of the unchanged area of the real-world, we perform photometric transform by style transfer using CycleGAN~\cite{zhu2017unpaired}.

After that, the generated pair image as shown in ~\figref{fig:fig2} is input to the U-Net-based original source image reconstructor during training. 
For our purpose, to train the original source image reconstructor with high reconstruction loss for the region where the change has occurred, we need to pay attention to the channel information in the original source image and pay attention to the spatial information in the photometric transformed image.
To achieve this purpose, we applied spatial attention to the photometric transform image and channel attention to the original source image using the CBAM~\cite{woo2018cbam}.

Despite these efforts, the original source image reconstructor has a problem of overfitting the original source image too easily during training.
Therefore, to prevent overfitting, we made a discriminator and conducted adversarial training with the image reconstructor.

\subsection{Photometric Transformation}
\label{sec:photo}
The purpose of photometric transform for training CDRL is to create a natural style change while maintaining the structure like an actual unchanged bi-temporal pair image as a single-temporal single source image.
In order to achieve this purpose, we adopted CycleGan~\cite{zhu2017unpaired}, which receives unpaired images and changes the style while maintaining the structure.
In the existing CycleGan, when there are two domains $\{ x_{1}, x_{2}, ..., x_{n}\} \in X$ and $\{ y_{1}, y_{2}, ..., y_{n}\} \in Y$, it receives two samples $x_{i}$ and $y_{j}$ and is trained to optimize the parameters of two mapping functions $G: X \rightarrow  Y$, $F: Y \rightarrow  X.$
However, since we need to perform unpaired style transfer in one domain, we train a function that maps two randomly selected samples $x^{t1}_{i} \in X^{t1}$ and $x^{t2}_{i} \in X^{t2}$ in one domain $X$.
Therefore, when there is discriminator $D_{t2}$ for mapping function $G: X^{t1} \rightarrow X^{t2}$ and discriminator $D_{t1}$ for $F: X^{t2} \rightarrow X^{t1}$, our objective function is as follows:
\begin{equation}
\begin{aligned}
  L(G,F,D_{t1},D_{t2}) = L_{GAN}(G,D_{t2}, X^{t1}, X^{t2}) \\
  + L_{GAN}(F,D_{t1}, X^{t2}, X^{t1}) + \lambda L_{cyc}(G,F),
  \label{eq:cycle}
\end{aligned}
\end{equation}
where $\lambda$ controls the relative importance of the two objectives.
%


\subsection{Pair Image-based Source Image Reconstructor}
\label{sec:model}
The pair image-based Source Image reconstructor $R(.)$ is trained to reconstruct $X^{t1}$ by receiving the pseudo unchanged pair image $X^{t1}$, $X^{t2}$ previously created in ~\sref{sec:photo} as an input.
To achieve this purpose, the pair image-based source image reconstructor consists of a shared encoder and a decoder that concats and fuses each feature map of the pair image output from the encoder.
$R$ is trained to optimize the objective function as follows:
\begin{equation}
\begin{aligned}
  L_{mae}(R) = MAE(R(X^{t1}, X^{t2}),X^{t1}),
  \label{eq:recon}
\end{aligned}
\end{equation}
where ${MAE}$ is the mean absolute error between the reconstructed image and the source image.

\begin{figure}[h!]
  \centering
  \includegraphics[width=\linewidth]{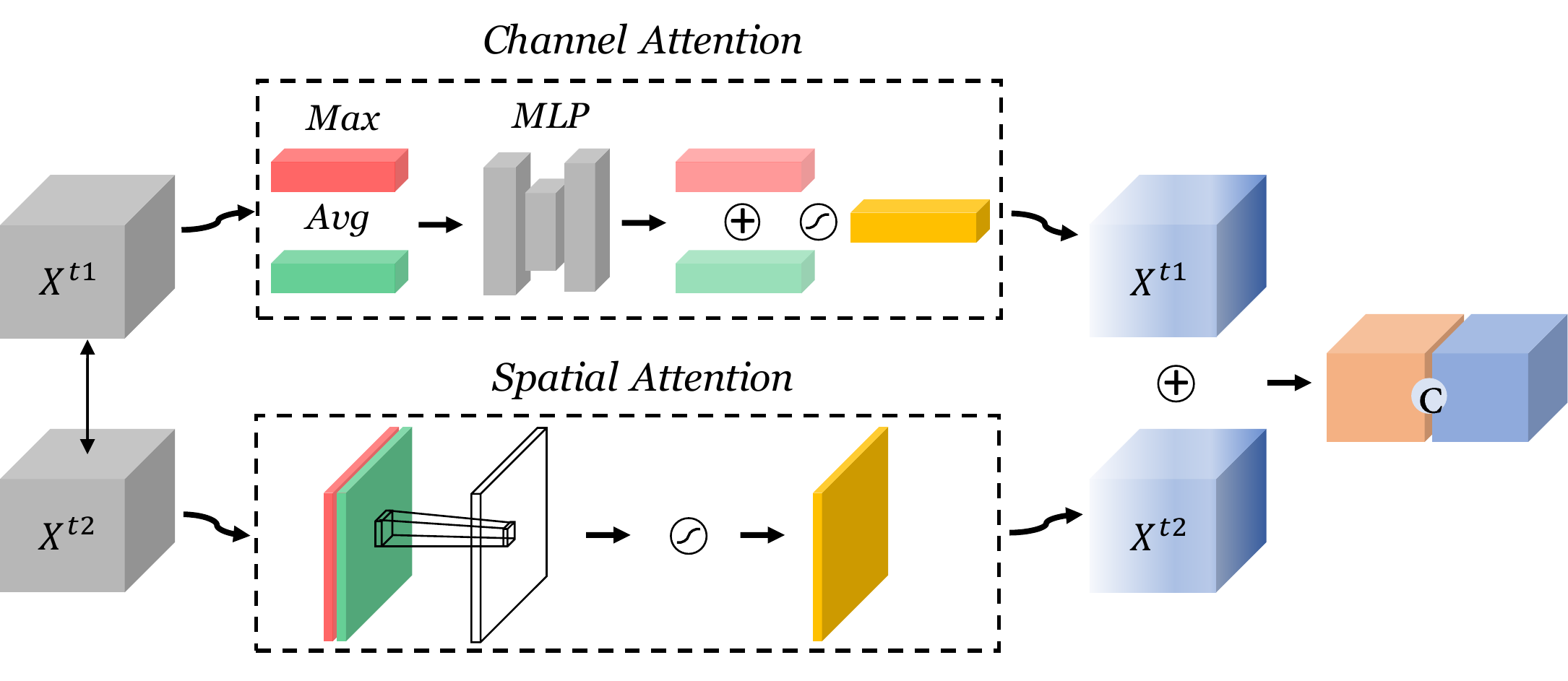}
  \caption{Our proposed pseudo-pair CBAM structure. Channels are applied to the $X^{t1}$ image, and spatial attention is applied to the $X^{t2}$ image}
  \label{fig:fig9}
\end{figure}

A source image reconstructor trained only on pseudo unchanged pair images during training should have a high reconstruction loss when a changed pair image is received during inference.
However, if the source image reconstructor reconstructs by relying only on the structure information of the source image regardless of the photometrically transformed image, the reconstruction loss is low even when a changed pair image is input.
To alleviate this problem, we modified the CBAM structure to perform spatial attention on the photometric transformed image and channel attention on the original source image.
Through this process, the source image reconstructor is trained by paying attention to the structure information of the photometric transformed image and paying attention to the style information in the source image.
~\figureref{fig:fig9} shows our attention structure modified from the CBAM structure. As shown in the figure, channel attention is performed on the $X^{t1}$ image and spatial attention is performed on the $X^{t2}$ image, which is then added and concatenated to train.

\subsection{GAN for detailed structure reconstruction}
\label{sec:gan}
As in the ~\cite{isola2017image} study, if only the $MLE$ loss is used, the reconstruction image does not reconstruct the structure well and is blurry. If the reconstruction result is blurry, the performance of the CDRL is degraded because it is insensitive to structure changes. Therefore, we use GAN as in \cite{isola2017image} to solve this problem. So, given the discriminator $D_{r}$, the objective function of the discriminator is:
\begin{equation}
\begin{aligned}
  L_{gan}(R, D_{r}, X^{t1}, X^{t2}) = log(D_{r}(X^{t1})) \\ 
  + log(1-D_{r}(R(X^{t1}, X^{t2}))),
  \label{eq:dis}
\end{aligned}
\end{equation}
where $R$ tries to reconstruct images $R(X^{t1}, X^{t2})$ that look similar to images from $X^{t1}$ , while $D_{r}$ aims to distinguish between translated samples $R(X^{t1}, X^{t2})$ and original source image $X^{t1}_{i}$. $R$ aims to minimize this objective against an adversary $D_{r}$ that tries to maximize it, i.e., $min_{R} max_{D_{r}} L(R, D_{r} , X^{t1}, X^{t2} )$.

The final objective function of the source image reconstructor that combines the GAN loss and the MAE loss is
\begin{equation}
\begin{aligned}
  L_{total} = L_{gan} + \lambda L_{mae}
  \label{eq:dis}
\end{aligned}
\end{equation}
where $\lambda$ controls the relative importance of the two objectives. We use $\lambda$ as 100 in all experiments.
\section{Experiments}
In this section, CDRL is evaluated on two HSR remote sensing change detection datasets.
Section ~\sref{sec:setting} describes the experimental setting in detail, and section ~\sref{sec:toy} describes the loss analysis experimental results in detail.
Also, in ~\sref{sec:pixel}, pixel level change detection, which is the same as the existing change detection experimental settings, and in ~\sref{sec:patch}, we describe the patch level change detection results suitable for our proposed change detector application situation.
Finally, in ~\sref{sec:qual}, qualitative results, and in ~\sref{sec:able}, the results of the ablation study are described in detail.

\subsection{Experimental Setting}
\label{sec:setting}
\paragraph{Datasets} Two HSR remote sensing change detection datasets were used to train and evaluate the performance of object change detection.

\begin{itemize}
    \item \textbf{LEVIR-CD}~\cite{chen2020spatial}\textbf{.} The LEVIR-CD dataset contains 637 bi-temporal pairs of HSR images and 31,333 change labels on building instances. Each image has a spatial size of 1,024 $\times$ 1,024 pixels with a spatial resolution of 0.5 m. The change labels provide information about the construction of new buildings and the disappearance of existing buildings. This dataset provides an official split of 445 training, 128 validation, and 64 test pairs. The evaluation results are computed in the test pair set.
\end{itemize}

\begin{itemize}
    \item \textbf{WHU building change detection}~\cite{ji2018fully}\textbf{.} The WHU dataset has one pair of aerial images of size 15,354 $\times$ 32,507 pixels obtained in 2012 and 2016 at the same area. It provides 12,796 and 16,077 building instances labels, respectively, and changed labels across the pair. We will use the change labels later to evaluate change detectors. Training, validation, and test sets are given specific areas containing 4,736, 1,036, and 2,416 tiles respectively.
\end{itemize}
\paragraph{Implementation details} 
CycleGAN was used to generate pseudo unchanged pairs for training of CDRL. The two datasets $X^{t1}$ and $X^{t2}$ of CycleGAN are randomly divided into datasets $X$. All implementation details strictly follow the official CycleGAN code\footnote{https://github.com/junyanz/pytorch-CycleGAN-and-pix2pix}.
For the data augmentation, RandomRotate90, HorizontalFlip, VerticalFlip, Transpose, RandomBrightnessContrast, and Sharpen of albumentations~\footnote{https://github.com/albumentations-team/albumentations} were used, and the probability $p$ of almost all applications is $0.3$.


We trained the source image reconstructor using the Adam optimizer with beta values equal to (0.5, 0.999). The learning rate set to 0.0002 and batch size of 1 to train the model. evaluation is using both the LEVIR-CD and WHU dataset validation (or test) sets.  
Our all models are implemented on PyTorch and trained using a single NVIDIA Quadro RTX 8000 GPU. 
The detailed structure of the network will be available soon on our project page.
\paragraph{Evaluation Metrics}

\begin{itemize}
    \item \textbf{Pixel Level Change Detection.} We use the common metrics in pixel-by-pixel binary classification tasks and object change detection tasks: intersection over union (IoU), recall, precision score. Because our goal is also to classify whether it has changed or not at the pixel level.
\end{itemize}

\begin{itemize}
    \item \textbf{Patch Level Change Detection.} We used classification AUC for patch level change detection. Note that if all of the output mask values of the change detector are 0, it is set to unchanged, if at least one is 1, it is set to change
\end{itemize}

\begin{figure}[t!]
  \centering
  \includegraphics[width=\linewidth]{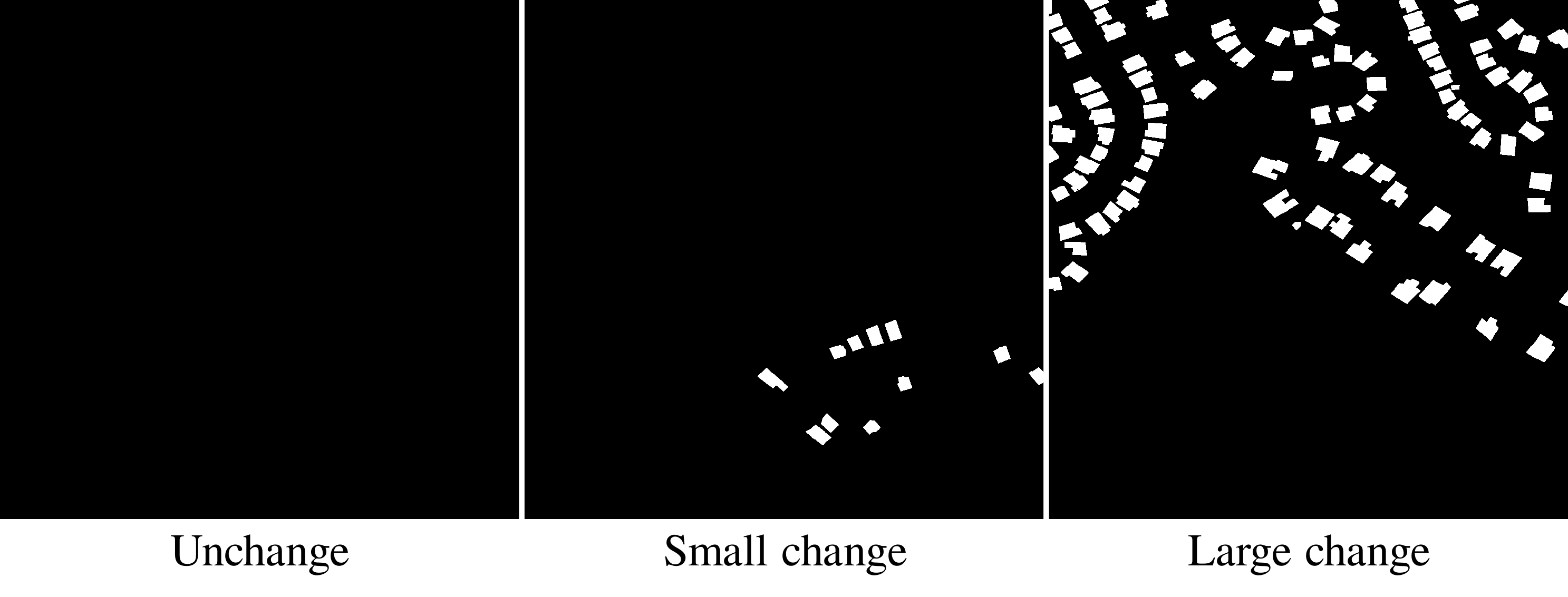}
  \caption{Sample images divided into unchange, small change, and large change for reconstruction loss analysis.}
  \label{fig:fig6}
\end{figure}
\begin{figure*}[t!]
  \centering
  \includegraphics[width=\linewidth]{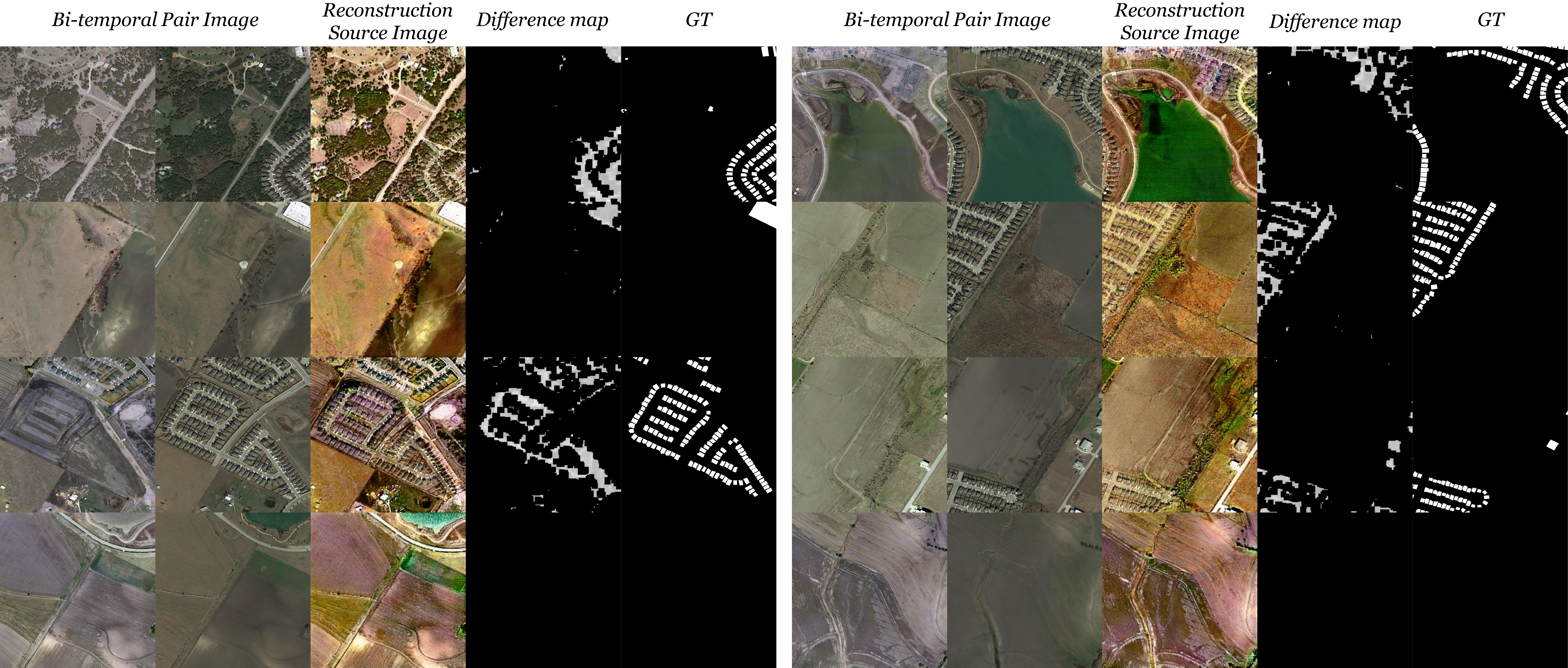}
  \caption{Qualitative analysis of CDRL. The top 3 lines are the qualitative results of CDRL in the region where the change has occurred, and the bottom line is the qualitative result of CDRL in the unchanged pair. It can be seen that CDRL localizes the area where the change occurred.}
  \label{fig:fig3}
\end{figure*}
\subsection{Loss Analysis Results}
\label{sec:toy}
We planned a loss analysis experiment to check whether the source image reconstructor has a high reconstruction loss in the part where the structure is changed a lot.

We divided the dataset into unchange, small change (the changed part is less than 30\% of the total image), and large change as shown in ~\figref{fig:fig6}.
When the dataset was split based on these criteria, the LEVIR-CD dataset was split into 8 unchanged pairs, 35 small change pairs, and 21 large change pairs.
The WHU dataset was split into 377 unchange pairs, 145 small change pairs, and 138 large change pairs.

\begin{table}
  \centering
  \resizebox{0.47\textwidth}{!}{
  \begin{tabular}{c | c | c c c}
    \toprule
    Method & Dataset &  \multicolumn{3}{c}{Reconstruction Loss} \\
     &  & Un & Small & Large \\
    \midrule
    CDRL (Aug) & LEVIR-CD& 24.79 & 36.81 & 44.91 \\
    CDRL (Pseudo Unchange Pair) & LEVIR-CD & 14.45 & 34.74 & 43.49 \\
    CDRL (Aug+Pseudo Unchange Pair) & LEVIR-CD& 10.15 & 31.41 & 38.53 \\
    \hline
    CDRL (Aug) & WHU& 35.95 & 41.11 & 50.89 \\
    CDRL (Pseudo Unchange Pair) & WHU & 22.89 & 39.37 & 49.46 \\
    CDRL (Aug+Pseudo Unchange Pair) & WHU& 17.20 & 38.03 & 47.65 \\
    \bottomrule
  \end{tabular}}
  \caption{Loss analysis results of CDRL in LEVIR-CD dataset and WHU dataset.}
  \label{tab:tab1}
\end{table}

~\tableref{tab:tab1} shows the results of loss analysis of CDRL in the LEVIR-CD test dataset and the WHU test dataset.
As shown in the table, the loss of the unchanged pair was the lowest in both the LEVIR-CD dataset and the WHU dataset, and the loss of the large change pair was the highest.
These experimental results indicate that the source image reconstructor is not good at reconstructing the source image when a pair with a large change in structure is input during the test because only pseudo unchanged pairs were input during training, as we intended.
Also, the fact that the loss was low in the unchanged pair indicates that our pseudo unchanged pair was generated at a level similar to that of the actual unchanged pair.

However, if the source image reconstructor works perfectly as we intended, the reconstruction loss should be close to 0 when an unchanged pair is input.
As shown in ~\tableref{tab:tab1}, the unchanged pair showed the lowest loss, but the value was not small.
The reason for this is that the LEVIR-CD dataset and the WHU dataset are labeled as only changes in building, and in fact, the unchanged pair includes many structural changes such as land becoming lakes, roads that did not exist, and cars.
A more detailed analysis result is described with an example in ~\sref{sec:qual}.

\begin{figure*}[t!]
  \centering
  \includegraphics[width=\linewidth]{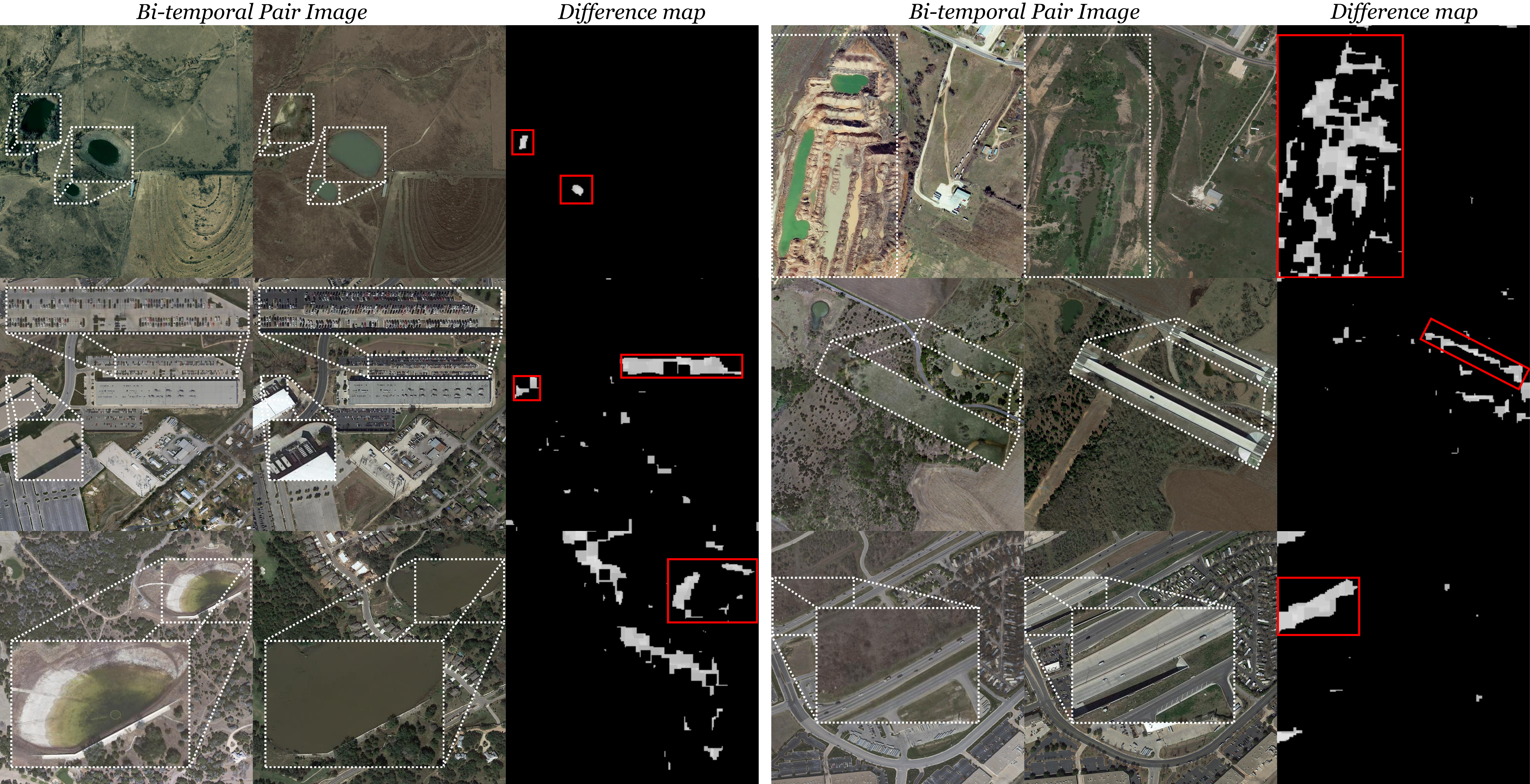}
  \caption{Qualitative analysis result of unchanged bi-temporal images for which CDRL showed high reconstruction loss. CDRL predicted that when roads, lakes, and cars were created, it was all change.
}
  \label{fig:fig7}
\end{figure*}

\subsection{Pixel Level Change Detection Results}
\label{sec:pixel}
We compared and analyzed the performance of CDRL with other UCD and SCD methods.
In order to compare our CDRL with the existing UCD methods, we reproduced all ~\cite{li2019unsupervised, de2019unsupervised, li2019unsupervised} methods and tested them in LEVIR-CD and WHU.
As shown in ~\tableref{tab:ex2}, the existing UCD methods have high recall values and low precision because prediction results are very noisy and vulnerable to small structural changes.
These experimental results indicate that our CDRL is robust to small structural changes and also to style changes.

However, when compared with BIT, which is a state-of-the-art supervised change detection, the performance of CDRL was low due to a large gap. This reason is mainly analyzed for two reasons.
First, unlike supervised change detections, CDRL does not learn the pixel-level change area explicitly, so it can localize only the approximate location. Therefore, compared to BIT, our CDRL has similar recall value but clearly lower precision.
Second, supervised change detection can be explicitly trained on information about change objects of interest, so it can explicitly learn that a car is created or a lake is changed to be unchanged, but our CDRL predicts that they are all changed.
%

\begin{table}[t!]
  \centering
  \resizebox{0.47\textwidth}{!}{
  \begin{tabular}{c| c | c | c c c}
    \toprule
    Method & Dataset & Supervision & Precision & Recall & IoU\\
    \midrule
    NMCD~\cite{li2019unsupervised} & LEVIR-CD & Unsup & 0.13 & 0.71 & 0.07\\
    UCNN~\cite{de2019unsupervised} & LEVIR-CD & Unsup & 0.16 & 0.79 & 0.09\\
    UCDGAN~\cite{li2019unsupervised} & LEVIR-CD & Unsup & 0.20 & 0.66 & 0.15\\
    \textbf{CDRL} & LEVIR-CD & Unsup & \textbf{0.63} & \textbf{0.92} & \textbf{0.59}\\
    BIT~\cite{chen2021remote} & LEVIR-CD & Sup & 0.89 & 0.89 & 0.80\\
    \hline
    NMCD~\cite{li2019unsupervised}  & WHU & Unsup & 0.07 & 0.96 & 0.03\\ 
    UCNN~\cite{de2019unsupervised}  & WHU & Unsup & 0.07 & 0.95 & 0.03 \\ 
    UCDGAN~\cite{li2019unsupervised}  & WHU & Unsup & 0.09 & 0.93 & 0.08\\ 
    \textbf{CDRL}  & WHU & Unsup & \textbf{0.52} & \textbf{0.93}  & \textbf{0.50}\\ 
    BIT~\cite{chen2021remote}  & WHU & Sup & 0.86 & 0.81 & 0.72 \\ 
    \bottomrule
  \end{tabular}}
  \caption{Quantitative comparison results of CDRL and UCD, SCD methods. Note that, since there are no post-processing implementation details of the existing UCD methods, it was not applied.}
  \label{tab:ex2}
\end{table}

\subsection{Patch Level Change Detection Results}
\label{sec:patch}
Many real-world applications that use change detectors do not rely solely on change detectors.
In these situations, the role of change detectors is to reduce human labor intensity by providing information on the patch or area where the change occurred among hundreds of patches.
Considering this application situation, we try to solve change detection with patch level classification.
\begin{table}[h!]
  \centering
  \begin{tabular}{c | c}
    \toprule
    Dataset & AUC \\
    \midrule
    LEVIR-CD & 83.52 \\ 
    WHU & 87.18 \\ 
    \bottomrule
  \end{tabular}
  \caption{Patch level binary classification results in the LEVIR-CD dataset and the WHU dataset.}
  \label{tab:auc}
\end{table}

~\tableref{tab:auc} shows the patch level change detection results for our CDRL in the LEVIR-CD dataset and the WHU dataset.
As shown in the table, high AUC was achieved in both datasets despite using only single-temporal single source images.
Note that CVA-based methods predict that there is a change in all patches because the output result is noisy.

\subsection{Qualitative Results}
\label{sec:qual}

CDRL was qualitatively analyzed on the LEVIR-CD dataset.
As shown in ~\figref{fig:fig3}, it can be seen that the CDRL detects the changed part well because the reconstruction loss is high in the part where the structure change is large.

Also, for the unchanged pair, even if the style change is large, since there is no structure change, it can be seen that the reconstruction loss is low and no change is predicted
Therefore, it is thought that CDRL will be useful in applications where it is important not to localize the exact location, but to know the approximate location or whether or not changes have occurred in units of patches.

~\figref{fig:fig7} shows the qualitative analysis results for samples with poor CDRL performance
As shown in the figure, since CDRL cannot designate a specific change object of interest, it predicts that a change has occurred when a car is created, the ground becomes a lake, or a lake becomes the ground.
Therefore, the reason for the low IoU in all our experiments is dominant for the above reasons.
\subsection{Ablation Study}
\label{sec:able}
In order to compare and analyze the effect of the attention module and adversarial training constituting the CDRL, we conducted an ablation study on the LEVIR-CD dataset.
\paragraph{Attention Modules }
~\tableref{tab:abl2} shows the patch level classification results of CDRL in the LEVIR-CD dataset according to the existence of channels attention and spatial attention.
As shown in the table, performance was higher with CBAM than without CBAM. Also, in our proposed CDRL, the pseudo-pair CBAM, which provides channels attention to the $X^{t1}$ image and spatial attention to the $X^{t2}$ image, has the highest performance.
\begin{table}[h!]
  \centering
  \begin{tabular}{c| c}
    \toprule
    CDRL & AUC \\
    \midrule
    w/o Attention & 77.38 \\ 
    w/ CBAM & 80.90 \\ 
    w/ Pseudo-pair CBAM & 83.52 \\ 
    \bottomrule
  \end{tabular}
  \caption{Patch level binary cleavage results of CDRL with or without attention module.}
  \label{tab:abl2}
\end{table}
\paragraph{Adversarial Training}

We designed an experiment to check whether adversarial training solves the blurry problem of reconstruction images like ~\cite{isola2017image} in CDRL.

%
\begin{figure}[h!]
  \centering
  \includegraphics[width=\linewidth]{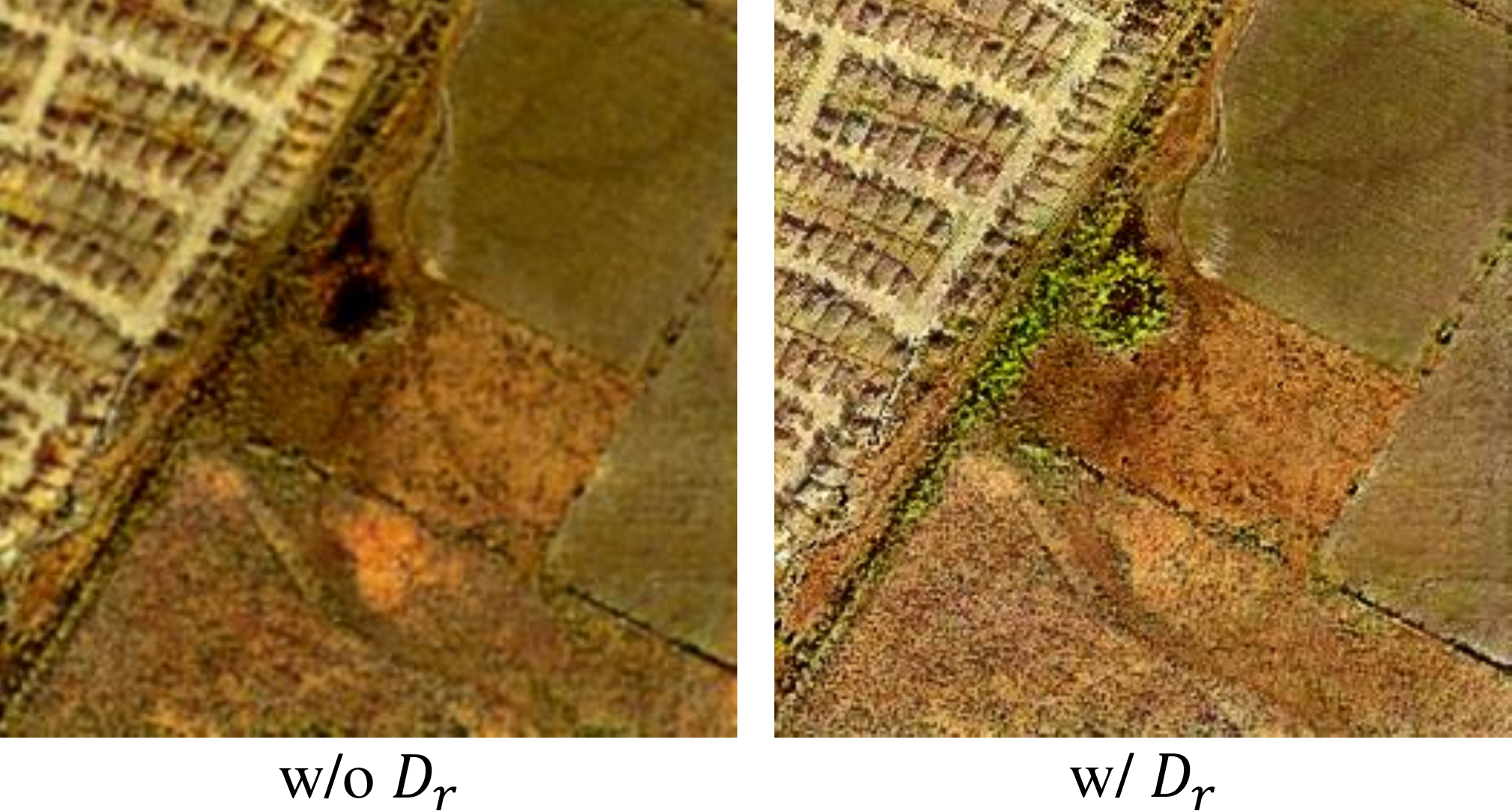}
  \caption{Qualitative comparison of reconstructed images with and without discriminator.}
  \label{fig:fig10}
\end{figure}

~\figref{fig:fig10} is a sample of the reconstructed image with and without discriminator.
As shown in the figure, it can be seen that the reconstructed image is blurred when there is no discriminator. On the other hand, if there is a discriminator, it can be seen that the boundary is reconstructed more clearly. 
These results indicate that adversarial training is also effective in CDRL.

~\tableref{tab:abl1} shows the results of measuring the AUC of CDRL with and without discriminator on the LEVIR-CD dataset. As shown in the table, it can be seen that higher AUC was achieved with the discriminator.
These experimental results show that the sensitivity of the CDRL to the structure of the pair image, like our design, helps the performance.

\section{Discussion and Future work}

In our work, we performed change detection with source image reconstruction loss using only unlabeled single-temporal single source images.
However, the semantic change we are interested in can exist in a variety of ways, such as buildings, seasons, cars, and trees.
The prerequisite for CDRL is to make the reconstruction loss appear high in the part where the structure change occurs regardless of the style change.
However, the semantic change that we are interested in can be diverse, such as natural scenery, artificial objects, weather, and environmental changes.
Therefore, in future work, based on the fact that CDRL has a significant performance improvement in UCD, we plan to study semi-supervised change detection to efficiently detect changes of interest (change of specific object)
\begin{table}
  \centering
  \begin{tabular}{c| c}
    \toprule
    CDRL & AUC \\
    \midrule
    w/o discriminator & 69.09 \\ 
    w/ discriminator & 83.52 \\ 
    \bottomrule
  \end{tabular}
  \caption{Patch level binary cleavage results of CDRL with or without discriminator.}
  \label{tab:abl1}
\end{table}
\section{Conclusion}
In this paper, to solve the problem that it is difficult to construct a bi-temporal pair dataset containing semantic changes, we propose a CDRL that performs unsupervised change detection using only a single-temporal single source image.
In order to solve the unsupervised change detection problem as a reconstruction-based unsupervised anomaly detection problem, CDRL defined normal data as unchanged pairs and anomaly data as changed pairs.
After that, a change detector (reconstructor) that receives pair images was proposed.
We verified the CDRL on the WHU and LEVIR-CD datasets, and achieved significant performance despite unsupervised change detection using single-temporal single source images.
We hope that CDRL will be widely used in real-world scenarios where it is difficult to obtain labeled bi-temporal pair images.

\label{sec:con}

{\small
\bibliographystyle{ieee_fullname}
\bibliography{egbib}
}

\end{document}